\def\year{2021}\relax
\documentclass[letterpaper]{article} 
\usepackage{aaai21}  
\usepackage{times}  
\usepackage{helvet} 
\usepackage{courier}  
\usepackage[hyphens]{url}  
\usepackage{graphicx} 
\usepackage{natbib}  
\usepackage{caption} 
\usepackage{latexsym}
\usepackage{booktabs}
\usepackage{color}
\usepackage{amssymb}
\usepackage{amsmath, calc}
\usepackage{multirow}
\usepackage{graphicx}
\usepackage{makecell}
\usepackage{subfigure}
\usepackage{algorithm}
\usepackage[noend]{algcompatible}
\newcommand{\taboo}{Adversarial Taboo}
\newcommand{\tabincell}[2]{\begin{tabular}{@{}#1@{}}#2\end{tabular}}
\title{Adversarial Language Games \\ for Advanced Natural Language Intelligence}
\author{
    Yuan Yao\thanks{indicates equal contribution}, Haoxi Zhong$^{*}$, Zhengyan Zhang, Xu Han, Xiaozhi Wang, \\ Kai Zhang, Chaojun Xiao, Guoyang Zeng, Zhiyuan Liu\thanks{Corresponding author: Z.Liu(liuzy@tsinghua.edu.cn)}, Maosong Sun\\
}
\affiliations{
    Department of Computer Science and Technology\\
Institute for Artificial Intelligence, Tsinghua University, Beijing, China\\
Beijing National Research Center for Information Science and Technology, China \\
\texttt{\{yuan-yao18,zhonghx18\}@mails.tsinghua.edu.cn}
}

\begin{document}

\maketitle

\begin{abstract}
We study the problem of adversarial language games\footnote{By ``language games'', we mean games played with language, not the philosophical term by~\citet{wittgenstein1953philosophical}.}, in which multiple agents with conflicting goals compete with each other via natural language interactions. While adversarial language games are ubiquitous in human activities, little attention has been devoted to this field in natural language processing. In this work, we propose a challenging adversarial language game called \textit{Adversarial Taboo} as an example, in which an attacker and a defender compete around a target word. The attacker is tasked with inducing the defender to utter the target word invisible to the defender, while the defender is tasked with detecting the target word before being induced by the attacker. In Adversarial Taboo, a successful attacker and defender need to hide or infer the intention, and induce or defend during conversations. This requires several advanced language abilities, such as adversarial pragmatic reasoning and goal-oriented language interactions in open domain, which will facilitate many downstream NLP tasks. To instantiate the game, we create a game environment and a competition platform. Comprehensive experiments on several baseline attack and defense strategies show promising and interesting results, based on which we discuss some directions for future research. The code and datasets of this paper can be obtained from \url{https://github.com/thunlp/AdversarialTaboo}. 
\end{abstract}

\section{Introduction}

\begin{figure}[t]
    \centering
    \includegraphics[width=\columnwidth]{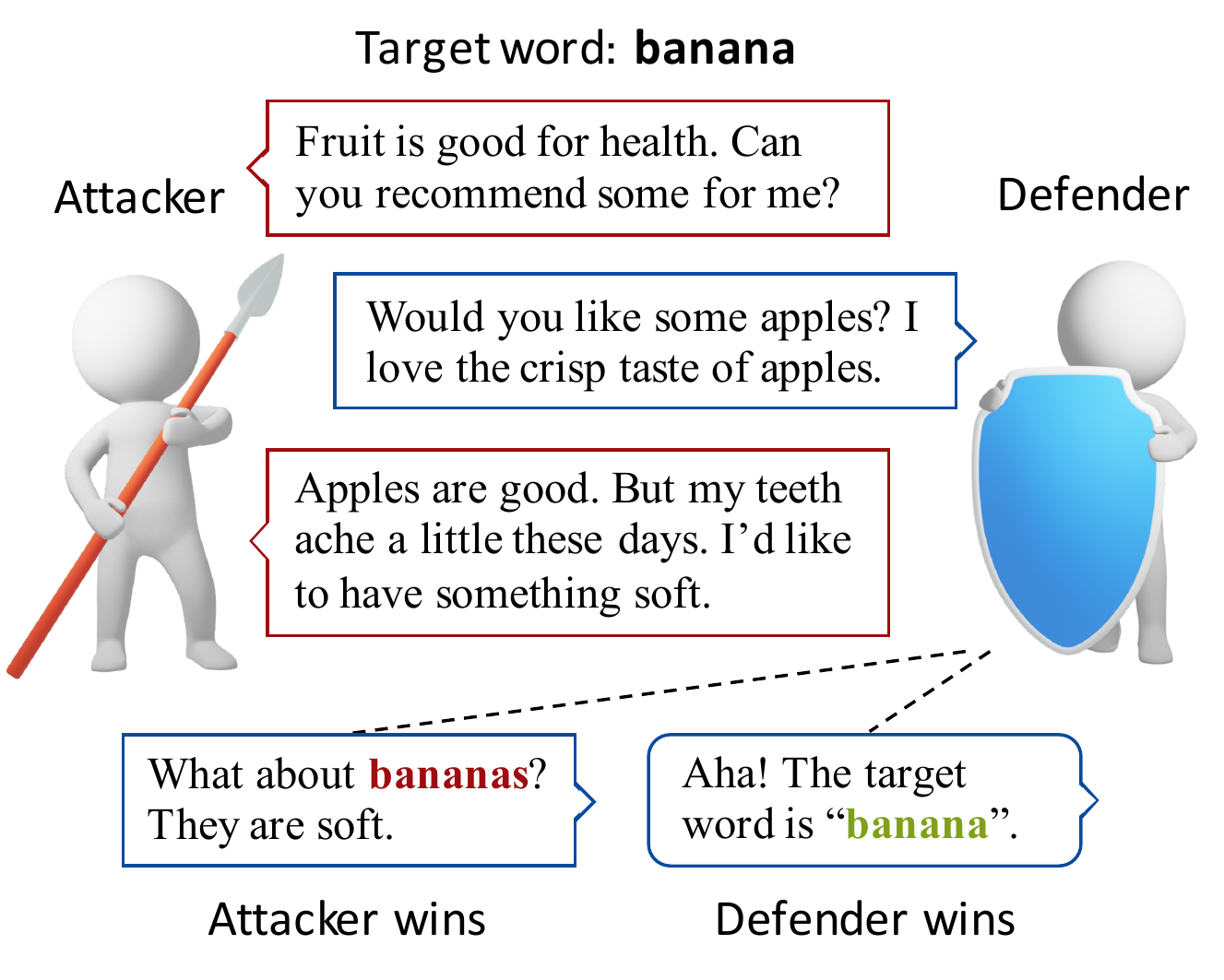}
    \caption{An example of Adversarial Taboo played by two human players, where the attacker and the defender compete through sequential language interactions. The target word ``banana'' is only visible to the attacker. Two possible cases of this game are shown. In the first case, the attacker wins since he/she successfully induced the defender to utter the target word. In the second case, the defender wins since he/she successfully inferred the target word of the attacker.}
    \label{fig:example}
\end{figure}

Natural language is inherently an interactive game between participants, which is ubiquitous in human activities such as discussion, debate, intention concealment and detection. Such context-related interactions are believed to play a central role in natural language mastery in the theory of both linguistics~\cite{mey2001pragmatics} and philosophy of language~\cite{wittgenstein1953philosophical,lewis1969convention}. High-quality goal-oriented natural language interactions, i.e., pragmatic interactions, generally require advanced language intelligence beyond syntax and semantics, and are particularly challenging due to the complexity, diversity and latent obscurity of natural language.

In the context of natural language processing (NLP), recent years have witnessed the success of deep learning on natural language understanding and generation. Language patterns learned from large-scale data lead to intelligent agents that can interact with humans with reasonable adequacy, fluency and diversity~\cite{radford2019language,brown2020language}. However, the intelligence of such agents is mainly confined to syntax and semantics, and devotes less attention to pragmatics~\cite{gao-etal-2018-neural}. Advanced language mastery (e.g., goal-oriented complex language skills and strategy usage in open domains) is still far from reach. It is believed that such advanced language intelligence can be better achieved through interactive language games~\cite{mikolov2016roadmap}.


Cooperation and adversary are both important in interactive language games. Previous work on cooperative language games studied cooperative pragmatic reasoning (i.e., infer the intention of the partner in contexts) and the emergence of language when agents share a common goal~\cite{strubend2017end,he2017learning,khani2018planning}. 
In comparison, adversarial language games require pragmatic reasoning in adversarial scenarios, and encourage agents with conflicting goals to proactively explore new complex language strategies (e.g., inducement with subtlety).
In a broader context, adversarial games have significantly promoted the development of many artificial intelligence areas such as board games \cite{campbell2002deep,silver2016mastering} and electronic sports games \cite{vinyals2017starcraft,jaderberg2019human}, and have enabled the emergence of complex strategies and superhuman proficiency in many cases. While some adversarial language games have been explored in persuasion~\cite{prakken2006formal} and negotiation~\cite{sadri2001dialogues,lewis-etal-2017-deal}, they can be (or need to be) simplified into formal language, where interactions are defined by specific rules on a finite set of atomic actions. Fewer efforts are devoted to adversarial games that need to be played in natural language. 


To this end, we propose a novel language game called \textit{Adversarial Taboo} as an example of adversarial natural language games, in which the attacker and the defender compete with each other through sequential natural language interactions. The goal of the attacker is to induce the defender to unconsciously utter a target word, which is given by the game system and invisible to the defender, and prevent the target word from being detected by the defender. Meanwhile, the defender aims to avoid the target word in utterances. The defender is also given one chance that can be used at any point to predict the target word. 

Figure \ref{fig:example} shows an example of Adversarial Taboo. The attacker is assigned with a target word ``banana'' by the judge system. In the first turn, the attacker asks for fruit recommendation, which is obscurely related to banana. Since the defender responds with ``apple'', in the second turn, the attacker continues to lead the topic more specifically to banana. The game can be terminated with two possible cases: (i) The defender says ``banana'' in his/her utterances, which leads to the win of the attacker. (ii) The defender successfully predicts the target word. If the game does not terminate within certain turns of interactions (e.g., $10$), the defender is forced to make predictions. 

Several complex language capabilities are required in Adversarial Taboo, including adversarial intention reading and concealment, inducement and inducement prevention. The attacker is required to obscure its intention (i.e., the target word) and subtly induce the defender, striking a balance between obscurity and inducement. A successful defender must balance between maintaining the semantic relevance of the response and preventing being induced, and at the same time, infer the attacking intention. Hence, mastering Adversarial Taboo leads to fine-grained language understanding, inference and generation, which is effective and convenient for language intelligence research. The game can also serve as a benchmark for language intelligence beyond syntax and semantics.~\footnote{Although syntax and semantics are also crucial to our game, we emphasize more on pragmatics, considering that plenty of works have focused on syntax and semantics.} We refer readers to the Outlook section for more discussions about research questions. 

Moreover, the language abilities required in \taboo\ can also facilitate many important real-world NLP tasks. For example, subtle intention inducement, or topical guidance, can be useful in therapeutic, educational and advertising conversations~\cite{tang-etal-2019-target}. Identifying (usually corrupted) user intentions is also crucial to customer service agents and search engines~\cite{yu2020few}. It's also desirable to build chatbots that are resistant to malicious inducement due to ethical concerns~\cite{srivastava2020personalized}.

To assess Adversarial Taboo, we propose attack and defense strategies that instantiate the required capabilities. We conduct comprehensive experiments including simulations between agents, and games between agents and human players. Experimental results show that simple attack and defense strategies can achieve promising and interesting results, while the proposed targeted improvements in strategies lead to alternate rises in the performance of attackers and defenders. 

Our contributions are summarized as follows: (i) We formulate a novel language game called Adversarial Taboo for advanced natural language intelligence. (ii) We propose several attack and defense strategies that instantiate the required capabilities, and conduct comprehensive experiments on the proposed game. (iii) Based on the experimental results and analysis, we discuss multiple directions for future research.

\section{Adversarial Taboo}
\label{sec:game}

\subsection{Task Definition}

In this section, we formalize the setting of \taboo. In the game, we denote the attacker as $A$ and the defender as $D$. We assume that $A$ will always speak first to lead the topic of conversations. Besides, we define $J$ as the judge system representing the rules of the game.

When the game begins, the judge system $J$ will first assign a target word $w$ to $A$. The game will last at most $T$ turns, and in every turn, $A$ and $D$ will take turns to utter a sentence.~\footnote{Each agent can utter multiple sentences. Here without losing generality, we discuss the scenario of one sentence.} Let's denote the sentence uttered by one player as $s$. Then $J$ will check the following: (1) The fluency and adequacy of $s$. To ensure the quality of the game, $J$ will check the syntactic correctness of $s$. Moreover, since the players are required to chat in the game, $J$ will check whether $s$ is relevant to the conversation context. (2) The outcome of the game. If $s$ is uttered by  $D$, $J$ checks whether $s$ contains the target word $w$. If $s$ contains $w$, then $A$ wins.\footnote{Note that here ``contain'' means that $s$ contains $w$ or any morphological variation of $w$. For example, the word ``bananas'' is a morphological variation of ``banana''.} 

$D$ is given one chance that can be used at any point to predict the target word. If $D$ correctly infers the target word, $D$ wins, otherwise the game continues. If the game does not end after $T$ turns, $D$ is forced to make predictions if he/she has not predicted the target word. If $D$ successfully infers the word, then $D$ wins, otherwise it is a tie.

\begin{figure*}[t]
\centering
\subfigure[Direct inquiries v.s. No defense.]{
\includegraphics[width=6cm]{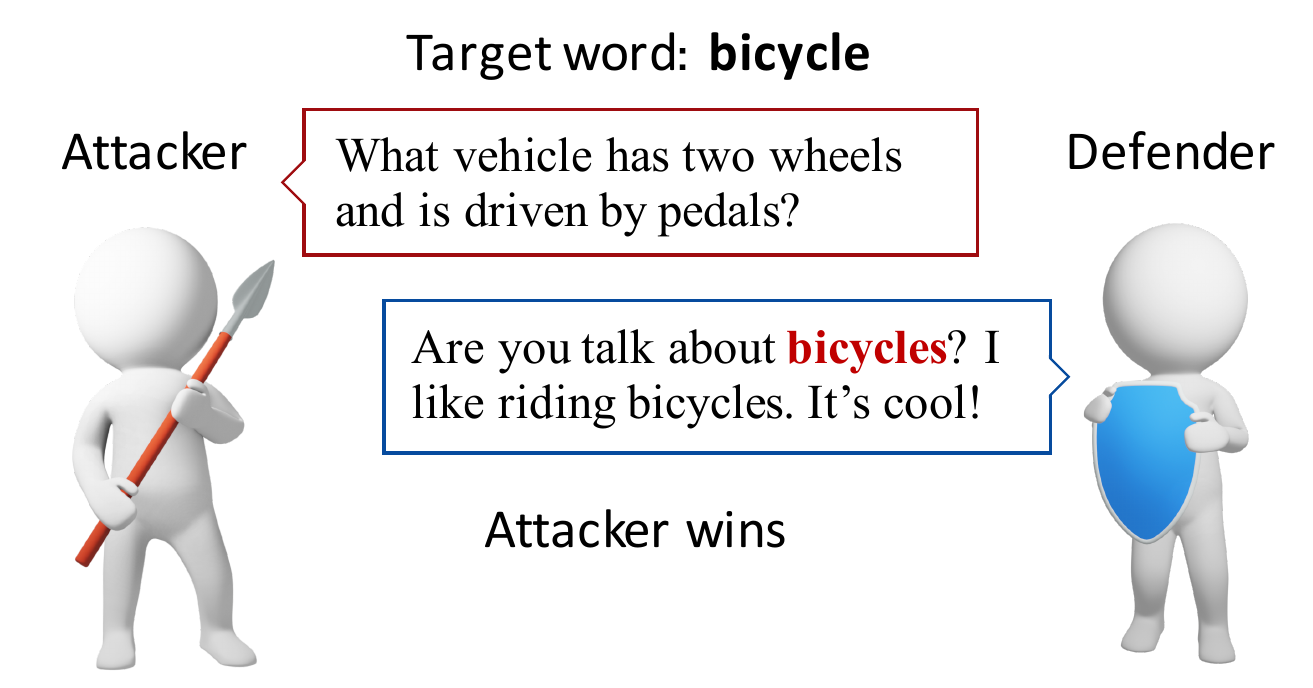}
}
\quad
\subfigure[Direct inquiries v.s. Intention detection.]{
\includegraphics[width=6cm]{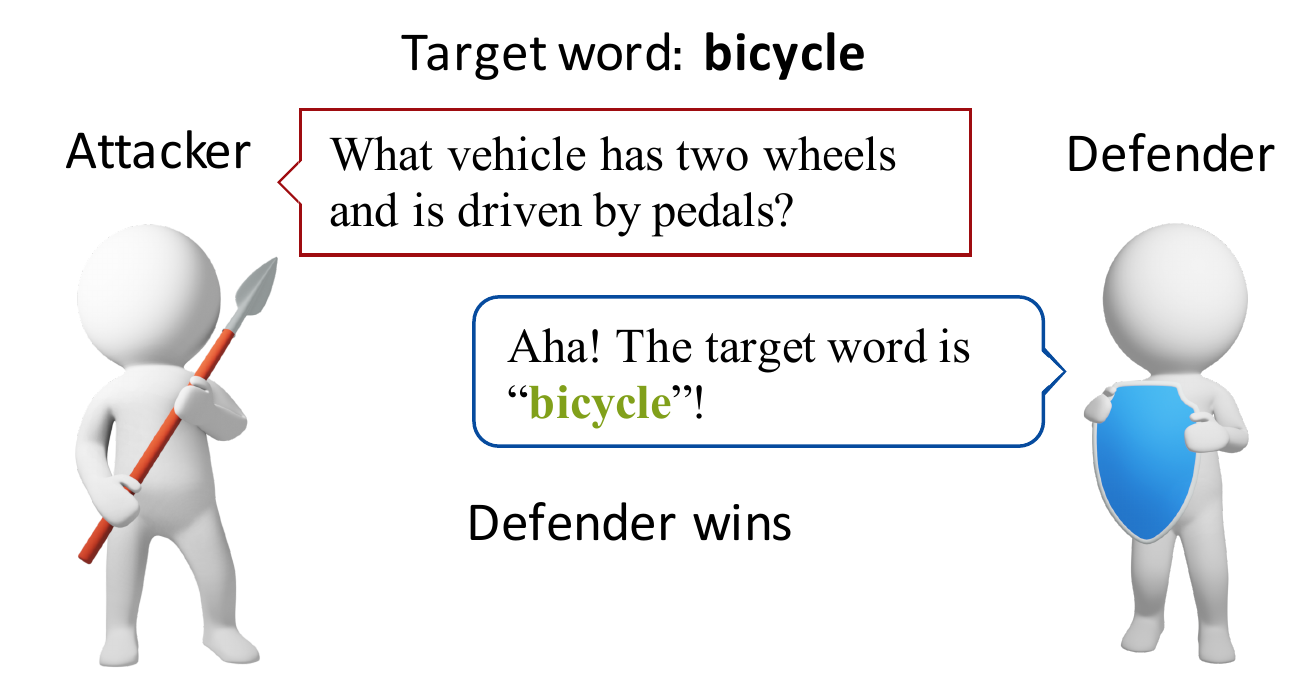}
}
\quad
\subfigure[Indirect inducement v.s. Intention detection.]{
\includegraphics[width=6cm]{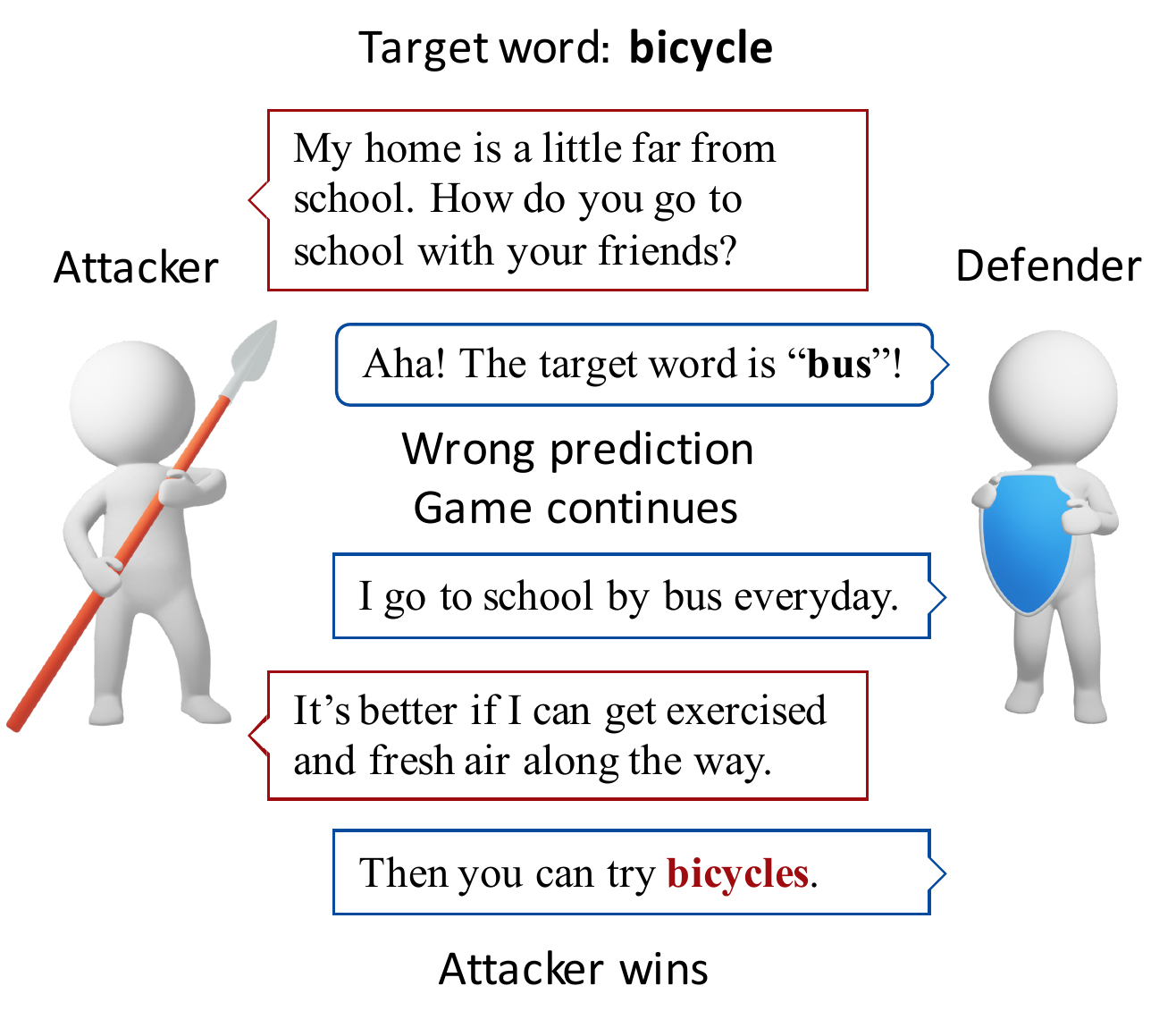}
}
\quad
\subfigure[\mbox{Indirect inducement v.s. Inducement prevention.}]{
\includegraphics[width=6cm]{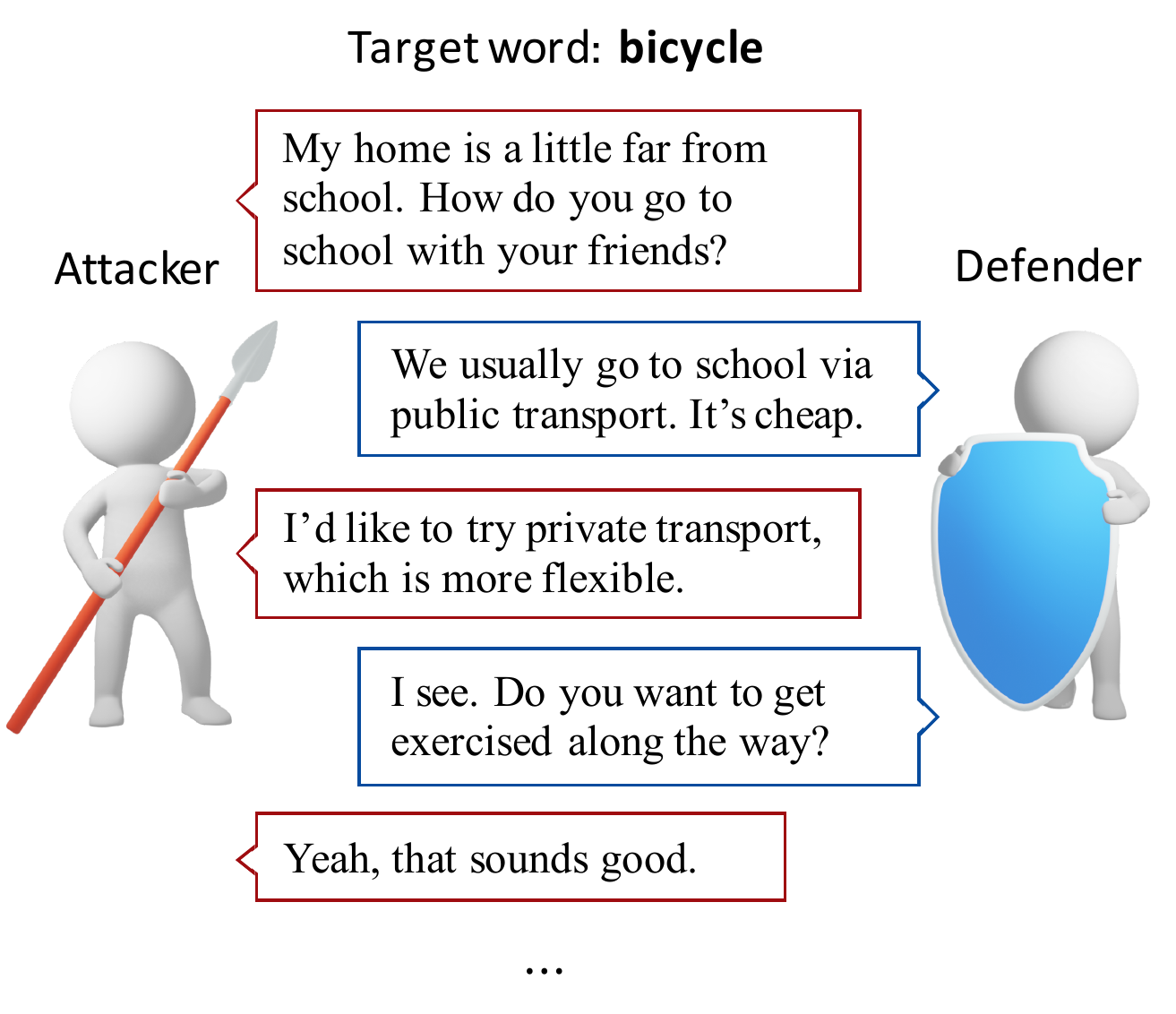}
}
\caption{Alternate improvements in attack and defense strategies with complex language skills that could emerge through co-adaptation in \taboo. (a) \textbf{Stage 1}: A defender without sense of defending will be successfully attacked by direct inquiries. (b) \textbf{Stage 2}: Direct inquiries will be easily defeated by a defender via intention detection. (c) \textbf{Stage 3}: The attacker needs to hide its intention to prevent being detected by the defender. (d) \textbf{Stage 4}: Both the attacker and defender need to be cautious with utterances when trying to achieve their goals.}
\label{fig:evolution}
\end{figure*}

\subsection{Competition Simulation}
In this section, we empirically instantiate and assess the game of \taboo. We simulate the competition on our game platform, including baselines and iterative improvements of agents (strategies) on the leaderboard. The simulated competitions include open question answering (OpenQA) based~\cite{chen2017reading} and chatbot-based~\cite{ritter2011data} models as backbones. OpenQA-based setting is a simplified scenario of the game where it is convenient to explicitly investigate each required capability, while chatbot-based setting is closer to real-world scenarios. 

In each simulation setting, we propose several attack and defense baseline strategies that attempt to instantiate the capabilities required in \taboo, and analyze the strategies based on the performance in competition. The instantiated language capabilities and game scenarios in the simulation can also be well aligned with many real-world NLP application tasks. 

\textbf{Judge System.}
We fine-tune a pre-trained language model using GPT-2~\cite{radford2019language} to check the fluency of single sentences, and BERT~\cite{devlin2019bert} to check the relevance of a response and a post. We refer the readers to appendix for more experiment details.

\textbf{Target Words Selection.}
The target words are nouns with high frequency selected from background corpus. Specifically, we select $563$ target words from English Wikipedia\footnote{\url{https://en.wikipedia.org}} articles for OpenQA-based simulation, and $567$ target words from Reddit conversation dataset~\cite{zhou2018commonsense} for chatbot-based experiment. For each target word, we simulate $5$ rounds of games, where each round consists of at most $10$ turns of interactions (i.e., $T=10$).


\begin{table*}[t]
    \centering
    \small
    \begin{tabular}{ll||cccc|cccc}
    \toprule
    \multirow{2}{*}{Attack} & \multirow{2}{*}{Defense} & \multicolumn{4}{c|}{DocQA }& \multicolumn{4}{c}{BERT} \\
     & & Attacker & Defender & Tie & \# Turns & Attacker & Defender & Tie & \# Turns\\
    \midrule
    Direct  & N/A    & 99.5 &N/A & 0.5 & 1.94 & 99.3 &N/A & 0.7 & 1.97\\
    Direct  & Detection  & 39.7 & 59.9 & 0.4 & 1.91 & 43.9 & 55.4 & 0.7 & 1.99\\
    Indirect  & Detection  & 70.8 & 26.1 & 3.1 & 3.64 & 70.7 & 25.9 & 3.4 & 3.61 \\
    Indirect  & Prevention & 55.7 & 28.5 & 15.8 & 4.87 & 58.8 & 30.2 & 11.0 & 4.43 \\
    \bottomrule
    \end{tabular}
    \caption{Competition simulation results on OpenQA-based models. The winning rate (\%), tie rate (\%) and the average turns of the game are reported.  N/A denotes that no strategy is adopted or the result is not applicable.}
    \label{Table:openQA_reuslt}
\end{table*}

\begin{table*}[t]
    \centering
    \small
    \begin{tabular}{ll||cccc|cccc}
    \toprule
    \multirow{2}{*}{Attack} & \multirow{2}{*}{Defense} & \multicolumn{4}{c|}{ConceptFlow}& \multicolumn{4}{c}{DialoGPT} \\
     & & Attacker & Defender & Tie & \# Turns & Attacker & Defender & Tie & \# Turns\\
    \midrule
    Topic Leading & \multirow{4}{*}{N/A} & 6.2 &N/A & 93.8 & 9.45 & 4.5 &N/A & 95.5 & 9.62 \\
    Golden Trigger  &    & 37.0 & N/A& 63.0 & 8.04 & 29.3 & N/A& 70.7 & 8.51  \\
    Neural-based     &   & 29.5 &N/A & 70.5 & 8.18 & 29.6 & N/A& 70.4 & 8.36 \\
    API-based     &   & 50.9 & N/A& 49.1 & 6.67 & 16.3 &N/A & 83.7 & 9.16  \\
    \midrule
    Golden Trigger & \multirow{3}{*}{Defense}    & 32.9 & 5.6 & 61.5 & 7.87 & 28.8 & 1.6 & 69.6 & 8.49 \\
    Neural-based & & 23.3 & 7.4 & 69.3 & 7.99 & 27.9 & 1.7 & 70.4 & 8.36  \\
    API-based &  & 38.2 & 14.6 & 47.2 & 6.31 & 15.0 & 0.7 & 84.3 & 9.14 \\
    \bottomrule
    \end{tabular}
    \caption{Competition simulation results on chatbot-based models.} 
    \label{Table:chatbot_attack_reuslt}
\end{table*}

\section{OpenQA-based Simulation}
OpenQA-based simulation is a simplified version of the game, where the attacker asks questions about target words, and the defender returns answer spans consisting of several words based on background corpus. Despite the simplicity, an advantage of OpenQA-based simulation is that each capability required in \taboo\ can be explicitly implemented and assessed. 

Specifically, the attacker is instantiated by a neural question generation model~\cite{lewis-etal-2019-unsupervised} that generates questions based on target words and sentences from Wikipedia. The defender is instantiated by an OpenQA system with two components: a paragraph retriever that first retrieves $3$ most relevant paragraphs from Wikipedia corpus using BM25, and a machine reading comprehension model that identifies answer spans from the retrieved paragraphs. We conduct experiments on two state-of-the-art reading comprehension models, including DocQA~\cite{clark-gardner-2018-simple} and a BERT-based model~\cite{devlin2019bert}. We simulate the iterative improvements of attack and defense strategies in four stages, as illustrated in Figure~\ref{fig:evolution}.

\textbf{Stage 1: Direct inquiries v.s. No defense.} The attacker $A$ asks questions on the target word in a straightforward way and the defender $D$ answers the question directly. 
In each turn, $A$ randomly selects a sentence that contains the target word from Wikipedia, and then generates a question on the target word. 
Stage 1 tests the basic communication ability of players, which is a prerequisite of \taboo.

\textbf{Stage 2: Direct inquiries v.s. Intention detection.} Now, $D$ will predict the target word based on the confidence of an answer span: $\text{conf}(a)= \exp(S_{start}(a) + S_{end}(a))$, where $S_{start}(\cdot)$ and $S_{end}(\cdot)$ denote the logits of the start and end token of an answer respectively. $D$ will use the opportunity to infer the target word when the confidence of an answer is greater than a threshold, otherwise it will utter the answer with top confidence.

\textbf{Stage 3: Indirect inducement v.s. Intention detection.} Since the intention of direct inquiries will be easily detected by $D$, $A$ needs to obscure its intention during inducement. A simple and intuitive strategy is to hide the intention (i.e., target word) in relevant topics. Specifically, $A$ extends a one-hop graph centered on the target word in the ConceptNet~\cite{speer2017conceptnet}, a large-scale commonsense knowledge graph. The concepts in the graph are linked by commonsense relations. $A$ first chooses a concept from the graph by random walks biased towards the target word, and then asks questions on the concept.

\textbf{Stage 4: Indirect inducement v.s. Inducement prevention.} Since the intention of $A$ cannot be directly detected from a single inquiry, $D$ needs to be cautious with the utterances, and infers the target word from multiple interactions. Specifically, given an inquiry, $D$ predicts a list of answers ranked by confidence. To prevent being induced, instead of always uttering the top-ranking answer $a_1$, $D$ utters the second-ranking answer $a_2$ that does not contain $a_1$, when $a_2$ is relevant enough to the question according to the confidence. $D$ keeps track of top predictions $a_1$ in different turns, and will use the opportunity to infer the target word according to the accumulated confidence of an answer.

\section{Chatbot-based Simulation}

In comparison to OpenQA-based experiments, posts and responses in chatbot-based simulation are in free-form, which is closer to real-world scenarios. We instantiate $D$ with two state-of-the-art generative chatbots: (1) DialoGPT~\cite{zhang2019dialogpt}, which adopts $345$M pretrained GPT-2~\cite{radford2019language} as backbone, and uses maximum mutual information reranking~\cite{li-etal-2016-diversity} to encourage the diversity of the response. (2) ConceptFlow~\cite{zhang2019grounded}, a chatbot enhanced with commonsense knowledge. ConceptFlow extends a graph in ConceptNet from concepts in the post, and mimics the conversation topic flow with graph attention for response generation. In our experiments, $A$ and $D$ are trained (or fine-tuned) on two disjoint dataset split from the Reddit dataset, ensuring that the training data of $D$ is invisible to $A$. 

It is non-trivial to integrate each capability required in \taboo{} into a generative chatbot. 
As an alternative, we can add some smart strategies to the chatbot models to guide them to play \taboo{}. 
This simulation can also be aligned with (1) topical guidance in therapeutic and educational conversations, and (2) attacks on publicly available chatbot service to produce (potentially unethical) target utterances. 
In our experiments, we assume that $A$ is a retrieval-based chatbot that selects posts from the Reddit dataset according to the attack strategy, while $D$ is a generative chatbot instantiated by DialoGPT or ConceptFlow.

\subsection{Strategies of Attackers}
We present four possible strategies for chatbot attackers.

\textbf{Strategy 1: Topic Leading Attack.} The simplest strategy of $A$ is selecting the initial post from the Reddit dataset, such that the golden response of the post in Reddit contains the target word. This strategy can lead conversation topic in the initial turn. 

\textbf{Strategy 2: Golden Trigger Attack.} Besides leading the topic at the beginning of the topic, $A$ can try to lead the topic every time in the game. $A$ can select the post such that its golden response contains the target word. The strategy can try to lead $D$ to utter the word in every turn.


\textbf{Strategy 3: Neural-based Attack.}
If $D$'s answer varies in the same context, then Strategy 2 will be less effective. As an alternative way, when the domain of $D$'s training corpus is available, predicting the target word in response can be formulated as a sentence classification problem.  
Specifically, we adopt BERT to obtain the distribution of whether the response contains the target word.
Then candidate posts from Reddit are selected according to the score predicted by the classifier. 

\textbf{Strategy 4: API-based Attack.}
In real-world scenarios, chatbots are usually exposed to public access via Application Programming Interface (API). Thus although the model architecture and training data are invisible to the public, the behavior of the chatbot is available to attackers through API access, which can potentially be used for more efficient targeted attacks. In our experiment, we implement the strategy by changing the training data for the classifier in Strategy 3 to be an alternative way of attacking on API. More specifically, the golden responses are replaced with responses inferred by the defender. API-based attack is also a kind of black-box attack, in the sense that the model architecture of $D$ is unknown to $A$.

\subsection{Strategies of Defenders}
Besides attack strategies, we also design strategies for defenders. In chatbot-based simulation, we wish to integrate intention detection into chatbots without adding new components. To this end, ConceptFlow can utilize its concept graph by choosing the concept with the highest attention score. For DialoGPT, since concepts in utterances are not explicitly modeled, the target word is detected based on the decoding score of each noun word in response generation.

\section{Simulation Results}

In this section, we present simulation results on OpenQA-based and chatbot-based agents.

\subsection{OpenQA-based Simulation Results}
The simulation results on OpenQA-based game, including the winning rate of each player, the tie rate and the average number of turns per round, are shown in Table~\ref{Table:openQA_reuslt}. We observe that the simulation results are consistent between the two OpenQA backbones.

(1) In stage 1, we observe a high winning rate of $A$ and a low average number of turns per round, which shows that the two players can communicate in reasonable quality. It also indicates that $D$ without a sense of defending is vulnerable even to direct inducement. 

(2) In stage 2, the winning rate of $D$ improves significantly, with a notable advantage over $A$, indicating that the intention of straightforward inquiries can be easily detected by $D$ using only one chance of prediction.

(3) In stage 3, $A$ dominates the game again by concealing the intention in relevant topics, which makes it difficult for $D$ to successfully predict the target word. We also observe an increase in the tie rate, since more turns are spent by $A$ in concealing the intention, leading to a decrease in inducement for the target word. 

\begin{figure}[t]
    \centering
    \includegraphics[width=0.85\columnwidth]{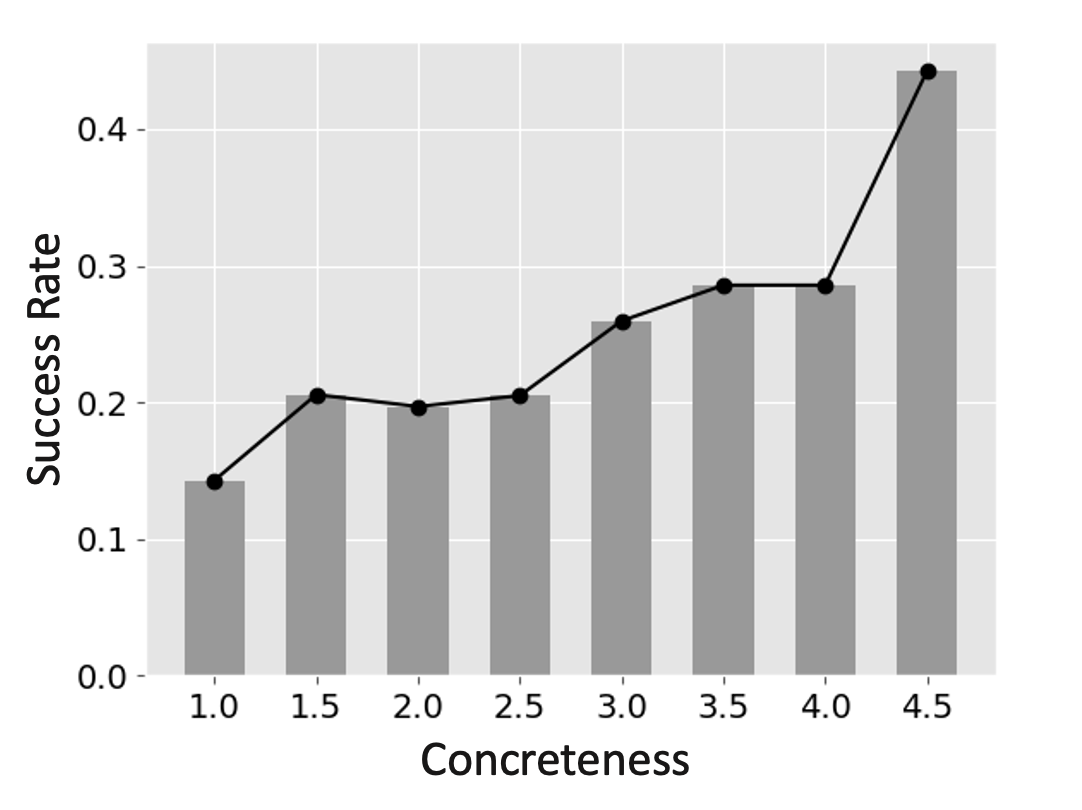}
    \caption{The correlation between the attack success rate and concreteness of words.}
    \label{fig:concreteness}
\end{figure}

(4) In stage 4, $D$ infers the intention of $A$ and prevents being induced. Although the tie rate and the winning rate of $D$ increase, $A$ still has an absolute advantage. This indicates that it is challenging to avoid being induced by a concealed intention while keeping the relevance of the response, and also challenging to infer the concealed intention from multiple turns of conversation, which requires more complex and strong strategies. We leave it for future work.

\subsection{Chatbot-based Simulation Results}

The chatbot-based simulation results are reported in Table~\ref{Table:chatbot_attack_reuslt},~\footnote{The perplexity of DialoGPT and ConceptFlow is $21.3$ and $34.4$ for reference. Note that the perplexities of the two models are not directly comparable due to different vocabularies.} from which we have the following observations:

(1)  When $A$ leads the topic, the attack success rate is low even if the initial post is highly relevant to the target word. It indicates that chatbots without any strategies can hardly play our game.

(2) When $A$ is guided with strategies not only at the beginning of the game, we observe a significant increase in the attack success rate. In particular, neural-based strategy achieves competitive results, which shows that the golden response is not necessarily a perfect indicator for attack due to the diversity of responses. We note that API-based attack achieves a high success rate in ConceptFlow, but a lower success rate in DialoGPT. The reason is that generative models pre-trained on large-scale data from different domains usually generate more diverse responses, making the response patterns hard to be captured by API-based strategy that is learned from a single domain. 

(3) We note that DialoGPT achieves a lower success rate in predicting the target word than ConceptFlow. This is because simply using the word-level language decoder for target word inference is sub-optimal, and will distract the inference of target words when the responses are diverse. In comparison, ConceptFlow explicitly models concepts in conversation development on sentence-level, which is more suitable for intention detection.

\begin{table}[t]
    \centering
    \small
    \renewcommand\arraystretch{1.2}
    \begin{tabular}{l p{.82\linewidth}}
    \toprule
    & \textbf{Target Word:} \textbf{{\color{red}\underline{Comedy}}}\\
    \midrule
    \textbf{A:} &  I really hope this {\color{red}\textit{film}} becomes a financial success. 
    It'd be nice if an actor was rewarded for putting so
    much faith and effort into a {\color{red}\textit{comic}} book character.\\
     
    \textbf{J:} & The defender predicts the word ``overrun". The prediction is wrong. Game continues. \\
    \textbf{D:} & I'd argue that the cost overrun on set prevent this 
    from being big success. We now have other {\color{blue}\textit{audience}} on the {\color{blue}\textit{screen}}. \\
    \textbf{A:} & I thought they scrapped doing a {\color{red}\textit{series}} because the {\color{red}\textit{movie}} flopped? \\
    \textbf{D:} & It didn't flop though, there literally just wasn't 
    that part of the entire {\color{blue}\textit{series}} and the third one that 
    opened with the compound {\color{blue}\textit{ending}} split up.\\
    \textbf{A:} & It was a great {\color{red}\textit{movie}} and an interesting concept 
    but I just don't find it as {\color{red}\textit{funny}} as people make out. \\ 
    \textbf{D:} & Agree. This has the mentality that {\color{red}\textbf{\underline{comedy}}} movies invented. \\
    \textbf{J:} & The attacker wins. \\

    \bottomrule
    \end{tabular}

    \caption{A game played by an attacker (\textbf{A}) and a defender (\textbf{D}), and supervised by the judge system (\textbf{J}). Concepts related to the target word mentioned by \textbf{A} and \textbf{D} are highlighted in {\color{red}\textit{red}} and {\color{blue}\textit{blue}} respectively.}
    \label{table:case}
\end{table}

\textbf{Vulnerability of Target Words.} 
Based on the simulation result, we investigate what kind of target words are more vulnerable to attack in generative chatbots. Specifically, we study the correlation between the attack success rate and the concreteness of words. The concreteness measures the perceptibility of the concept that a word presents (e.g., \textit{bicycle} is more concrete than \textit{intelligence}). We adopt the concreteness annotation from~\citet{brysbaert2014concreteness}. Figure~\ref{fig:concreteness} shows that concrete words are more vulnerable to attack, and inducing highly abstract words is challenging.

\textbf{Case Study.}
Table~\ref{table:case} shows a game played by two agents, including an attacker with neural-based strategy, and a defender instantiated with DialoGPT. We observe that, to a certain extent, the attacker can guide the topic around the target while concealing the intention. We expect future research on \taboo{} will lead to more complex language skills and strategies.

\section{Human Evaluation}
Besides simulations between agents, we also instantiate \taboo{} between agents and human players. Specifically, $A$ is an agent equipped with golden-trigger attack strategy, while $D$ is a human player. The human players are native English speakers. We conduct $50$ games in two settings respectively: (1) \textit{Control Setting}, where human players are not aware that they are playing our game; (2) \textit{Experimental Setting}, where human players are aware of the game and rules. The results are reported in Table~\ref{table:human performance}, from which we observe that: (1) The winning rate of the attacker is high in both settings, which indicates that humans are vulnerable to attack in \taboo, even if they are aware of participating in the game. (2) Compared to control setting, the overall performance of human defenders improves in experimental setting, which shows the effectiveness of human defense strategies. However, the game remains challenging for human defenders. The overall results show consistency with the simulation results between agents. We refer readers to the appendix for game examples of human evaluation.

\begin{table}[t]
\centering
\small

\begin{tabular}{l|c c c c}
\toprule

Setting & Attacker & Defender & Tie & \# Turns\\
\midrule

 Control  & 60.0 & N/A & 40.0 & 7.04 \\ 
 Experimental  & 46.0 & 10.0 & 44.0 & 7.24 \\

\bottomrule
\end{tabular}
 \caption{Evaluation results of games between an agent attacker and a human defender. The winning rate (\%), tie rate (\%) and the average turns of the game are reported.}
\label{table:human performance}
\end{table}

\begin{table*}
    \centering
    \small
    \begin{tabular}{lcccccc}
    \toprule
    Language Game & Cooperative & Adversarial & Asymmetric & \tabincell{c}{Formal\\Language} & \tabincell{c}{Natural\\Language} & \tabincell{c}{Open-domain\\Knowledge}\\
    \midrule
    Referential Games & \checkmark & & \checkmark & \checkmark & \checkmark &\\
    Taboo & \checkmark & & \checkmark & \checkmark & \checkmark & \checkmark\\
    Persuasion & \checkmark & \checkmark & & \checkmark & &\\
    Negotiation & & \checkmark & \checkmark  & \checkmark & \checkmark &\\
    Avalon & \checkmark & \checkmark & \checkmark & \checkmark & \checkmark &\\
    Werewolf & \checkmark & \checkmark & \checkmark & \checkmark & \checkmark & \\
    Who is the Spy &  & \checkmark & \checkmark & \checkmark & \checkmark & \checkmark\\
    Adversarial Taboo & \checkmark & \checkmark & \checkmark & & \checkmark & \checkmark\\
    \bottomrule
    \end{tabular}
    \caption{Different language games and their properties, including whether the language game is cooperative and adversarial (or both), whether the information of the players is asymmetric, whether the game can be simplified into formal language, whether the interactions can be in the form of natural language, and whether open-domain knowledge helps (or is required by) the game.}
    \label{Table:language_games}
\end{table*}

\section{Related Work}

\textbf{Language Games and Pragmatics.}
We compare different language games in Table~\ref{Table:language_games}, including the language games that have been investigated, and the ones that are unexplored but promising for future research (e.g., \taboo\ and Who is the Spy). Referential Games~\cite{lewis1969convention} can be generalized to a broad family of cooperative games, where one agent aims to select a specific object from candidates based on the unidirectional descriptions from a partner~\cite{xu2010inference,havrylov2017emergence,bouchacourt2018agents,kharitonov2019egg} or bidirectional interactions between two agents ~\cite{wang2016learning,strubend2017end,liu2017iterative,wei2018airdialogue,shah2018bootstrapping}, or two agents each with incomplete private information communicate to achieve a common goal~\cite{vogel2013implicatures,he2017learning,khani2018planning}. Cooperative language games can also be used to annotate data~\cite{von2006games}. Language games are related to pragmatics, which studies language meaning in the interactional context~\cite{mey2001pragmatics}, and plays an important role in linguistics and language teaching~\cite{kasper2001pragmatics}. There are also some works~\cite{smith2013learning,monroe2015learning,hawkins2015you,andreas-klein-2016-reasoning} that study the pragmatic reasoning ability of NLP models with pragmatics language games~\cite{krauss1964changes,clark1986referring,potts2012goal} and pragmatics theories such as the speech-act theory ~\cite{searle1980speech} and Rational Speech Act framework~\cite{golland-etal-2010-game,goodman2016pragmatic}. However, existing pragmatics games that require natural language interactions mainly focus on cooperation rather than competition. In \taboo, agents with conflicting goals compete through natural language to win the game. We refer readers to the appendix for a more detailed discussion.



\textbf{Adversarial Attack.}
Existing adversarial attack methods mainly focus on statically attacking models with corrupted semantics and enhancing the model robustness~\cite{jia-liang-2017-adversarial,cheng-etal-2019-robust}. Adversarial Taboo challenges models with dynamic adversarial interactions, and is expected to enhance the adversarial language skills. \citet{cheng2019evaluating} study adversarial learning in negotiation dialogues~\cite{lewis-etal-2017-deal}, where agents divide items based on conversations that can be simplified into formal language
~\cite{sadri2001dialogues}. \citet{tang-etal-2019-target} appoint an agent with the task of guiding the conversation topic to a target subject, where the inclusion of the target in either of the two participants' utterances will lead to the success of the task. In comparison, \taboo\ involves two different proactive roles in an adversarial language game, and requires multiple complex language skills in open domain. 

\textbf{Dialogue Systems} can be divided into goal-oriented systems and non-goal-oriented systems. Goal-oriented dialogue systems aim to assist users to accomplish certain tasks (e.g., booking restaurants)~\cite{goddeau1996form,williams2013dialog}, while non-goal-oriented dialogue systems (chatbots) generate natural responses in open domains by maximizing the likelihood of human responses~\cite{ritter2011data,li2016persona,serban2016building}. To better approximate the real-world goal of dialogue agents, recent years have witnessed a rising interest in developing dialogue systems through goal-oriented interactions~\cite{li2016deep,das2017learning,lewis-etal-2017-deal}. Adversarial Taboo takes the form of conversations, and we absorb many settings in dialogue systems to define our task.

\section{Outlook}

In this section, we discuss several promising directions for future research.

\subsection{Natural Language Capabilities Desired in \taboo}

We hope playing \taboo{} will help to develop and benchmark natural language capabilities as follows that are not yet well achieved by the current learning from static corpus paradigm. 

\textbf{Adversarial Pragmatic Reasoning.}
In cooperative language games, agents with private information communicate to achieve common goals. In many real-world scenarios, however, agents need to perform reasoning in adversarial contexts. Our game creates an adversarial context where agents with conflicting goals deliberately hide information and mislead the opponent for individual goals. 

\textbf{Goal-oriented Language Interaction in Open-domain.}
Natural language interactions are inherently goal-oriented, in the sense that humans interact via natural language to achieve certain (cooperative or adversarial) goals. \taboo{} tasks the agents with goals that require natural language interactions in open domain.

\textbf{Knowledge Enhanced Language Interaction.}
Human language generally involves a variety of knowledge, such as commonsense knowledge and world knowledge. In \taboo{}, knowledge utilization enables fine-grained indirect inducement and adversarial intention reasoning.

\textbf{Emergence of Language Skills via Co-evolution.}
Human language intelligence can proactively evolve through interactions. The competition pressure in \taboo\ also encourages the agents to explore new language skills, which will create new pressure for the opponent to adapt. Such competition and co-evolution paradigms have been shown promising in many artificial intelligence areas.

\subsection{Robust Judge System}
In our game, the judge system aims to prevent agents from generating unreadable or irrelevant sentences. However, the diverse and complex natural language interactions cannot be explicitly defined with a small set of rules, which makes the automatic evaluation particularly challenging. 
The evaluation has also received increasing attention from the dialogue community~\cite{liu2016not,ghazarian2019better}. Therefore, the research for a more robust judge system can remarkably benefit the evaluation of language generation. 

\section{Conclusion}
In this paper, we study the problem of adversarial language games, and propose \taboo\ as an example. We propose several attack and defense strategies, and conduct comprehensive experiments. Despite the promising experimental results, the proposed strategies are still simple and show large room for improvement. We expect investigating adversarial language games in NLP will both promote and benchmark the development of advanced natural language intelligence.

\section{Acknowledgement}
This work is supported by the National Key Research and Development Program of China (No. 2020AAA0106501), the National Natural Science Foundation of China (NSFC No. 61772302), and Beijing Academy of Artificial Intelligence (BAAI). The authors came up with the idea of \taboo{} when playing the game on Chang'an Avenue on September 2019.

\bibliography{aaai21}
\clearpage
\appendix

\section{Appendices}
\subsection{Language Games}
We compare several important properties of language games. (1) \textit{Cooperative} and \textit{Adversarial} indicates whether the players share the same utility. Note that a language game can be both cooperative (within groups) and adversarial (between groups). (2) \textit{Asymmetric} indicates whether the information and roles of players are asymmetric. Asymmetry in adversarial language games often leads to deception and interrogation. (3) \textit{Formal Language} denotes whether the interactions can be simplified into formal language, where interactions are defined by specific rules on a finite set of atomic actions. \textit{Natural Language} indicates whether the game can be played in natural language. Note that although many language games (e.g., Negotiation and Avalon) can take the form of natural language, they can be simplified into formal language. (4) \textit{Open-domain Knowledge} indicates whether open-domain knowledge helps (or is required by) the game. We believe the ability to incorporate and utilize open-domain knowledge is crucial to advanced natural language intelligence. 

We give a brief description of the language games in Table~\ref{Table:language_games}: (1) \textit{Taboo} is a variant of referential games, where the target object is a word. However, in this setting, players can utilize the shortcut to win the game easily. For example, if the target word is ``Memory'', the first player only needs to say ``Long short-term what?'' to finish the game. (2) \textit{Navigation} is a single-player game where an agent seeks to reach a location based on the textual descriptions from the game environment~\cite{narasimhan2015language}.  (3) \textit{Persuasion} is a game where multiple agents with conflicting opinions persuade each other~\cite{prakken2006formal} or an audience~\cite{amgoud2013axiomatic}. (4) \textit{Negotiation} requires agents to divide items with different values based on conversation~\cite{sadri2001dialogues,lewis-etal-2017-deal}. (5) \textit{Avalon} and \textit{Werewolf} are two popular role-playing language games, where the attackers seek to disrupt the defenders without being discovered, while the defenders need to identify the hidden attackers among them~\cite{AIwolf2017,ws-2019-international-ai}. (6) In \textit{Who Is the Spy}, one agent (the spy) is assigned with a word, while the rest agents, assigned with a different but similar word, aim to identify the spy through language interactions. (7) More generally, \textit{\taboo{}} can involve multiple groups of attackers and defenders. In such cases, players within each group cooperate with each other, while players in different groups compete with conflicting goals.

\subsection{Human Evaluation: Case Study}

We show examples of games between agent attackers and human defenders. Specifically, the defender is instantiated with a golden-trigger attack strategy. We show games under the control setting and the experimental setting in Table~\ref{table:control} and Table~\ref{table:experimental} respectively. From the results, we observe that:

(1) Humans can leverage open-domain knowledge, including commonsense knowledge and world knowledge for deep interactions in the game (e.g., understanding that \textit{Milwaukee} is a sports team instead of a location, and leveraging knowledge about related sports teams and players). However, the game is challenging even for human defenders. 

(2) Although the posts selected by the attacker are of reasonable fluency, the pragmatic rationale of the posts is less 

\begin{table}[t]
    \centering
    \small
    \renewcommand\arraystretch{1.2}
    \begin{tabular}{l p{.82\linewidth}}
    \toprule
    & \textbf{Target Word:} \textbf{{\color{red}\underline{Bull}}}\\
    \midrule
    \textbf{A:} &  I think {\color{red}\textit{Milwaukee}} is the only one we haven't beat yet that we've played. Still haven't played {\color{red}\textit{Washington}} though.\\
    
    \textbf{D:} & \vspace{-1em} {\color{blue}\textit{Milwaukee}} have the Greek freak so that is understandable. I think you would beat the {\color{blue}\textit{Wizards}} though. They have been poor lately.\\
    
    \textbf{A:} & The {\color{red}\textit{Bucks}}. Depending on when {\color{red}\textit{Rose}} gets back, we may want the extra time for him to shed the rust.\\ 
    
    \textbf{D:} & Yeah {\color{blue}\textit{Rose}} is rusty so he won’t be at his best ever again. He is just playing for fun at this stage. All he is trying to do is to avoid another {\color{blue}\textit{injury}}. \\
    
    \textbf{A:} & They don't need to free up cap space to max {\color{red}\textit{Jimmy}}, they have {\color{red}\textit{Bird Rights}}. And if this year has shown us anything, it's shown that you can never have too many big men.\\
    
    \textbf{D:} & \vspace{-1em} {\color{blue}\textit{Jimmy Butler}} is the perfect power forward for this {\color{red}\textbf{\underline{Bulls}}} team but he know they are underachievers. He might move to {\color{blue}\textit{Miami}} next. \\
    \textbf{J:} & The attacker wins. \\
    
    \bottomrule
    \end{tabular}

    \caption{Control setting: A game played by an agent attacker (\textbf{A}) and a human defender (\textbf{D}), and supervised by the judge system (\textbf{J}). Concepts related to the target word mentioned by \textbf{A} and \textbf{D} are highlighted in {\color{red}\textit{red}} and {\color{blue}\textit{blue}} respectively.}
    \label{table:control}
\end{table}
\hspace{-1em}favorable. This shows that more sophisticated strategies need to be investigated for better adversarial pragmatic modeling and planning. We leave it for future work.

\begin{table}[t]
    \centering
    \small
    \renewcommand\arraystretch{1.2}
    \begin{tabular}{l p{.82\linewidth}}
    \toprule
    & \textbf{Target Word:} \textbf{{\color{red}\underline{Mile}}}\\
    \midrule
    \textbf{A:} & With all this talk about turned down {\color{red}\textit{engines}}, has anyone heard of info on the {\color{red}\textit{Mercedes engines}}? Did they run with less {\color{red}\textit{power}}?\\
    
    \textbf{D:} & New regulations and {\color{blue}\textit{emissions}} requirements mean that all performance {\color{blue}\textit{engines}} will be tuned down over time. It is unfortunate but necessary.\\
    
    \textbf{A:} & Did not know that, we questioned if it was still the original {\color{red}\textit{engine}} and {\color{red}\textit{trans}}, up here in new England {\color{red}\textit{cars}} barely make it past 200, 000.\\ 
    
    \textbf{D:} & That is surprising since the climate doesn’t seem to be too harsh there. Maybe they are just poorly maintained or the chemicals used on the {\color{blue}\textit{roads}} are abrasive.\\
    
    \textbf{A:} & Well it's kind of true I guess. {\color{red}\textit{Parts}} do wear out but there's no point in not using them for fear of the inevitable.\\
    
    \textbf{J:} & The defender predicts the word ``vehicle". The prediction is wrong. Game continues. \\
    
    \textbf{D:} & That’s why there is a market for buying {\color{blue}\textit{replacement}} when things go wrong. We need to use these things anyway but wear and tear is part of the equation. \\
     & $\cdots$ \\
    \textbf{J:} & Game terminates with maximum turns in a tie. \\
    \bottomrule
    \end{tabular}

    \caption{Experimental setting: A game played by an agent attacker (\textbf{A}) and a human defender (\textbf{D}), and supervised by the judge system (\textbf{J}). Concepts related to the target mentioned by \textbf{A} and \textbf{D} are highlighted in {\color{red}\textit{red}} and {\color{blue}\textit{blue}} respectively.}
    \label{table:experimental}
\end{table}

\subsection{Implementation Details}

In this section, we provide implementation details about OpenQA-based simulation and chatbot-based simulations. 

\subsection{Target Word Selection}
We select the target words for OpenQA-based simulation from 2019-11-01 dump of English Wikipedia corpus that contains $1,200$M words. We select entities for target words, i.e., nouns with high frequency ($>8,000$) and high concreteness ($>4.5$). This results in $563$ target words for OpenQA-based simulation.

For chatbot-based simulation, we select the target words from Reddit dataset that contains $10$M words. We select nouns with high frequency ($> 200$). This results in $567$ target words for chatbot-based simulation.

\smallskip
\textbf{OpenQA-based Simulation.}
We provide details in each simulation stage: (1) In Stage 2, the confidence threshold of guessing the target word is $c_1$ ($10$ for BERT-based defender and $1e3$ for DocQA-based defender. (2) In Stage 3, the attacker extends a one-hop graph centered on the target word in the ConceptNet. The concepts in the graph are closely related to target words in semantics. We ensure this by only including concepts that are linked to target words by \texttt{Related\_To}, \texttt{Similar\_To}, \texttt{Is\_A} or \texttt{Has\_A} relations. The bias towards the target word in random walk is $0.6$, which indicates the transition probability to the target word from the current concept in each turn of conversation. (3) In Stage 4, the defender says the second-ranking answer if its confidence is greater than $c_2$ ($1.0$ for BERT-based defender and $1e5$ for DocQA-based defender) to keep the relevance of the answer. The defender will use the opportunity to predict the target word if the accumulated confidence of the top-ranking answer is greater than $c_3$ ($1e4$ for BERT-based defender and $1e6$ for DocQA-based defender), and the confidence in the current turn is greater than $c_1$.

\smallskip
\textbf{Chatbot-based Simulation.}
In chatbot-based simulation, we randomly split the Reddit dataset into two disjoint sets for the training of attacker and the defender respectively. We followed the implementation details in DialoGPT~\cite{radford2019language} and ConceptFlow~\cite{zhang2019grounded} to train the model: (1) For DialoGPT-based defender, we fine-tuned the pre-trained DialoGPT for $1$ epoch, with learning rate $1e$-$5$ and batch size $16$. The defender instantiated by DialoGPT will use the opportunity predict the top-ranking word if it is a noun and its decoding score is greater than $0.03$. (2) For ConceptFlow-based defender, we trained the model for $10$ epoch, with learning rate $1e$-$4$ and batch size $40$. We adopt $300$ dimensional GloVe~\cite{pennington2014glove} for word embedding initialization. Both the encoder and decoder are implemented by a $2$-layer GRU~\cite{GRU2014} with hidden state size $512$. The defender instantiated by ConceptFlow will predict the top-ranking concept if its score is greater than $0.1$. 

\smallskip
\textbf{Judge System.}
We adopt neural language models to measure the fluency and adequacy of sentences in experiments. 

(1) Fluency. Perplexity from language models is widely used to automatically evaluate the fluency of natural language generation~\cite{zhou2018commonsense,zhang2019grounded}. To measure the fluency, we fine-tune the $345$M GPT-2~\cite{radford2019language}, a state-of-the-art pre-trained language model, on the Reddit dataset using a language modeling objective. Given a sentence, we calculate the perplexity using GPT-2. A sentence is considered as syntactically correct if its perplexity is less than $35$, otherwise it is considered as invalid.

(2) Adequacy. It has been shown that BERT can well capture the relevance of sentences~\cite{devlin2019bert}. To measure whether the response is relevant to the post, we fine-tune the BERT-base model~\cite{devlin2019bert} on Reddit dataset using a classification objective. The positive post-response pairs come from post and golden response in a conversation. The negative post-response pair is obtained by randomly sampling sentences that are not the responses of a post. To prevent overly generic responses (which are usually more frequent than informative responses in conversation corpus), we subsampled the frequent responses in positive pairs to make the distribution of responses more balanced. The post and response tokens are first concatenated with a special token \texttt{[SEP]} separating the two segments, and then fed into the BERT model. The representation of \texttt{[CLS]} token is fed into a fully connected layer followed by sigmoid layer to obtain the relevance score. The classification loss is the cross-entropy loss. A response is considered as relevant to the post if the relevance score is higher than $0.4$, otherwise it is considered as irrelevant.

\end{document}